\documentstyle[]{lrec98}				

\title{Resources for Evaluation of Summarization Techniques}

\name{Judith L. Klavans* and Kathleen R. McKeown** and Min-Yen Kan** and Susan Lee***}
\address{
\begin{tabular}[t]{ccc}
Center for Research on Information Access* & Department of Computer Science** & Computer Sciences Division*** \\
Columbia University & Columbia University & University of California, Berkeley \\
New York, NY 10027 & New York, NY 10027 & Berkeley, CA 94720\\
\end{tabular}
}

\abstract{We report on two corpora to be used in the evaluation of
component systems for the tasks of (1) linear segmentation of text and (2)
summary-directed sentence extraction.  We present characteristics of the
corpora, methods used in the collection of user judgments, and an overview
of the application of the corpora to evaluating the component system.
Finally, we discuss the problems and issues with construction of the test
set which apply broadly to the construction of evaluation resources for
language technologies.}

\begin{document}

\maketitleabstract

\section{Application Context}

We report on two corpora to be used in the evaluation of component systems
for the tasks of (1) linear segmentation of text and (2) summary-directed
sentence extraction.

  Any development of a natural language processing (NLP) application
requires systematic testing and evaluation.  In the course of our ongoing
development of a robust, domain-independent summarization system at
Columbia University, we have followed this procedure of incremental testing
and evaluation\footnote{This research was supported by the NSF Award
9618797 in the Speech, Text, Image, and Multimedia in Language Technology
(STIMULATE) program entitled ``STIMULATE: Generating Coherent Summaries of
On-Line Documents: Combining Statistical and Symbolic Techniques" under the
direction of Kathleen R.  McKeown, Department of Computer Science and
Judith L. Klavans, Center for Research on Information Access''; Susan Lee
was supported at Columbia University by a summer internship under the
Computing Research Association (CRA) Distributed Mentor Project.}. However,
we found that the resources that were necessary for the evaluation of our
particular system components did not exist in the NLP community.  Thus, we
built a set of evaluation resources which we present in this paper.  Our
goal in this paper is to describe the resources and to discuss both
theoretical and practical issues that arise in the development of such
resources.  All evaluation resources are publicly available, and we welcome
collaboration on use and improvements.

The two resources discussed in this paper were utilized in the initial
evaluation of a text analysis module.  In the larger context, the analysis
module serves as the initial steps for a complete system for summarization
by analysis and reformulation, rather than solely by sentence extraction.
Analysis components provide strategic conceptual information in the form of
segments which are high in information content, and in which similar or
different; this information provides input to subsequent processing,
including reasoning about a single document or set of documents, followed
by summary generation using language generation techniques (McKeown and
Radev 1995, Radev and McKeown 1997).

\section{Description of Resources}

We detail these two evaluation corpora, both comprised of a corpus of human
judgments, fashioned to accurately test the two technologies currently
implemented in the text analysis module: namely, {\it linear segmentation}
of text and {\it sentence extraction}.

\subsection{Evaluation Resource for Segmentation}

The segmentation task is motivated by the observation that longer texts
benefit from automatic chunking of cohesive sections.  Even though
newspaper text appears to be segmented by paragraph and by headers, this
segmentation is often driven by arbitrary page layout and length
considerations rather than by discourse logic.  For other kinds of text,
such as transcripts, prior segmentation may not exist.  Thus, our goal
is to segment these texts by logical rhetorical considerations.

In this section, we discuss the development of the evaluation corpus for
the task of segmentation.  This task involves breaking input text into
segments that represent some meaningful grouping of contiguous portions
of the text.

In our formulation of the segmentation task, we examined the specifics of a
linear multi-paragraph segmentation of the input text, ``linear'' in that
we seek a sequential relation between the chunks, as opposed to
``hierarchical'' segmentation (Marcu 1997).  ``Multiple paragraph'' refers
to the size of the units to be grouped, as opposed to sentences or words.
We believe that this simple type of segmentation yields useful information for
summarization.  Within the context of the text analysis module,
segmentation is the first step in the identification of key areas of
documents.

Segmentation is followed by an identification component to label segments
according to function and importance within the document.  This labeling
then permits reasoning and filtering over labeled and ranked segments.  In
the current implementation, segments are labeled according to centrality
vis \`{a} vis the overall document.

\subsubsection{Segmentation Corpus}

To evaluate our segmentation algorithm's effectiveness, we needed to test
our algorithm on a varied set of articles.  We first utilized the publicly
available {\it Wall Street Journal} (WSJ) corpus provided by the Linguistic
Data Consortium.  Many of these articles are very short, i.e. 8 to 10
sentences, but segmentation is more meaningful in the context of longer
articles; thus, we screened for articles as close as possible to 50
sentences.  Additionally, we controlled our selection of articles for the
absence of section headers within the article itself, to guarantee that
articles were not written to fit section headers.  This is not to say that
an evaluation cannot be done with articles with headers, but rather that an
initial evaluation was performed without this complicating factor.

We arrived at a set of 15 newspaper articles from the WSJ corpus.  We
supplemented these by 5 articles from the on-line version of {\it The
Economist} magazine, following the same restrictions, to protect against
biasing our results to reflect WSJ style.  Although WSJ articles were
approximately 50 sentences in length; the {\it Economist} articles were
slightly longer, ranging from 50 to 75 sentences.  Average paragraph length
of the WSJ articles was 2 to 3 sentences, which is typical of newspaper
paragraphing, and 3 to 4 for the {\it Economist}.  Documents were domain
independent but genre specific in general terms, i.e. current events (any
topic) but journalistic writing, since this is the initial focus of our
summarization project.

\subsubsection{Task}

The goal of the task was to collect a set of user judgments on what is a
meaningful segment with the hypothesis that what users perceive to be a
meaningful unit will be useful data in evaluating the effectiveness of our
system.  The goal of our system is to identify segment boundaries and rank
according to meaningfulness.  The data could be used both to evaluate our
algorithm, or in later stages, as part of training data for supervised
learning.

To construct the evaluation corpus, subjects were asked to divide an
average of six selected articles into meaningful topical units at
paragraph boundaries.  The definition of segment was purposefully left
vague in order to assess the user's interpretation of the notion
``meaningful unit.''  Subjects were also encouraged to give subjective
strengths of the segments, if they wanted to.  Subjects were not told how
the segments would be used for later processing, nor informed of the number
of segment breaks to produce, and were given no further criteria for
choice.  Finally, subjects were not constrained by time restrictions;
however, subjects were given the tester's time estimate on task completion
time of 10 minutes per article (for both reading the article and
determining segment boundaries).  In total, 13 volunteers produced results,
all graduate students or people with advanced degrees.  A total of 19
articles were segmented by a minimum of four, and often five, subjects.
All 13 subjects segmented the one remaining single article.
 
\subsubsection{Analysis of Results of Human Segmentation}

The variation in segmentation style produced results ranging from very few
segments (1-2 per document) to over 15 for the longer documents.  As shown
in Table 1, the number of segments varied according to the length of the
article and specific article in question.  Most subjects completed the task
within the time we had initially estimated.

Subjects were found to be consistent in behavior: if they segmented one
article with fewer segments than the average, then the other articles
segmented by the subject were often also segmented with fewer breaks.  For
example, Subject 4 displays ``lumping'' behavior, whereas Subject 6 is a
``splitter''.  This points to an individual's notion of {\it granularity},
which is further discussed below in section 2.1.5.

\begin{table}[hbt]
\centering
\scriptsize
\begin{tabular}{|l|c|c|c|c|} 
\hline
\hline
& {\bf Article br7854} & {\bf Article am3332} & {\bf Article 0085} & {\bf Article 0090} \\
& \it{P} = 20 & \it{P} = 10 & \it{P} = 19 & \it{P} = 24 \\
\hline
\hline
Subject 1 & 6 & 4 & 7 & 7 \\
Subject 2 & 7 & 5 & 8 & 7 \\
Subject 3 & 10 & 4 & 11 & 9 \\
\hline
\end{tabular}
\end{table}

\begin{table}[hbt]
\centering
\small
\begin{tabular}{|l|c|c|c|} 
\hline
\hline
& {\bf Article fn6703} & {\bf Article mo0414} & {\bf Article 0071} \\
& \it{P} = 10 & \it{P} = 10 & \it{P} = 19 \\
\hline
\hline
Subject 4 & 2 & 2 & 5 \\
Subject 5 & 5 & 4 & 5 \\
Subject 6 & 9 & 7 & 16 \\
\hline

\end{tabular}
\label{t:table1}
\caption{Lumpers and Splitters problem on Segmentation Evaluation Corpora (where {\it P} = number of paragraph breaks in article)}
\end{table}

\subsubsection{Use}

To compile the gold standard we used majority opinion, as advocated by Gale
et al, 1992, i.e. if the majority indicated a break at the same spot, then
that location was deemed a segment boundary.  We compiled the judgments
into a database for use in optimal parameterization of a set of constraints
for weighting groups of lexical and phrasal term occurrences.  We
calculated a high level of interjudge reliability using Cochran's {\it Q},
significant to the 1\% level for all but 2 articles which were significant
to the 5\% level. See Kan et al, 1998 for further discussion of the use of
data in evaluating the segmentation algorithm.

\subsubsection{Issues}

The segmentation task is subject to interpretation, just like many natural
language tasks which involve drawing subjective boundaries.  Since the
directions were open-ended, responses can be divided into ``the lumpers''
and ``the splitters'', to use the terminology applied to lexicographers
when building dictionary definitions.  In the case of dictionary
construction, lumpers tend to write more terse, condensed definitions which
consist of several possible uses in one definition, whereas splitters will
divide definitions into a larger number of definitions, each of which may
cover only one aspect or one usage of the word.  For segmentation, the way
this tendency expressed itself is that the lumpers tended to mark very few
boundaries, whereas the splitters marked numerous boundaries.  In fact, as
mentioned above, some splitters marked over 15 segments for longer
articles, which is over 85\% of all possible paragraph breaks, on average.

For this reason, in determining what type of data to extract from the
evaluation corpus, we took only the majority segments for training and
testing; the result is that lumpers end up determining the majority.

\subsubsection{Future Work on the Segmentation Resource}

For future work, we would like to extend the resource to include a range of
genres (such as journal articles, documentation) as well as expand the
range of sources to include additional news articles (i.e. LDC's North
American News Text Corpus).  Also, we plan to extend our collection to
other languages since there is little research on applicability of general
techniques, such as segmentation based on terms and local maxima, across
languages for multilingual analysis tasks.  We are also considering
analyzing articles with section headers, to see whether they follow the
segment boundaries and if so, how they can be utilized for expanding an
evaluation resource.

In addition to expanding the corpus by genre, we also plan to collect
information for the segment labeling task.  In this stage, segments are
labeled for their function within the document.  In addition, this resource
will be useful in providing information on the function of the first (or
lead) segments.  In journalistic prose, the lead segment can often be used
as a summary due to stylized rules prescribing that the most important
information must be first.  However, the lead can also be an anecdotal
lead, i.e.  material that grabs the reader's attention and leads into the
article.  Thus, we plan to perform a formal analysis of how human subjects
characterize anecdotal leads.

\subsubsection{Availability}

The segmentation evaluation data is publicly available by request to the
third author.  Inquiries for the textual data that the evaluation
corpus is based on should be directed to the respective owners of the
materials.

\subsection{Evaluation Resource for Sentence Extraction}

In this section, we describe the collection of judgments to create the
evaluation resource used to test summary-directed sentence extraction.  One
method to produce a ``summary'' of a text is by performing sentence
extraction.  In this approach a small set of informative sentences are
chosen to represent the full text and presented to the user as a summary.
Although computationally appealing, this approach falls prey to at least
two major disadvantages: (1) missing key information and (2) disfluencies
in the extracted text.

Our approach takes steps to handle both of these problems and thus changes
what we mean by the sentence extraction task.  The majority of systems use
sentence extraction as a complete approach to summarization in that the
sentences extracted from the text are, in fact, the summary presented to
the user.  In the context of our system, we use the sentence extraction
component to choose a larger set of sentences than required for the
intended summary length.  All these sentences are then further analyzed for
the generation component that will synthesize only the key information
needed in a summary.  The synthesis procedure will eliminate some clauses
and possibly some whole sentences as well, resulting in a reformulated
summary of the intended length.  Thus, the goals of our ``sentence
extraction for generation'' task differ from ``sentence extraction as
summarization'' in that we seek high recall of key information.

\subsubsection{Extraction Corpus}

We used newswire text, available on the World-Wide Web from Reuters.  In
examining random articles available at the time of testing, we found that
the number of sentences per article were short: 18, on average.  Short
paragraphs were also a characteristic of the corpus, similar to the corpus
used for the segmentation evaluation: 1 to 3 sentences per paragraph on average.  These shorter texts enabled us to analyze more articles than in
the segmentation evaluation.  As a result we were able to double the number
of articles used for testing; we selected 40 articles, with titles, taken
from this on-line version.

\subsubsection{Task}

Na\"{\i}ve readers were asked to select sentences with high information
content.  Instructions were kept general, to let subjects form their own
interpretation of ``informativeness'', similar to the segmentation
experiment.  A minimum of one sentence was required, but no maximum number
was set.  All 15 subjects were volunteers, consisting of graduate students
and professors from different fields.  Subjects were grouped at random into
5 reading groups of 3 subjects each such that an evaluation based on
majority opinion would possible.  Each reading group analyzed 8 articles,
which covered the entire 40 article set.  Articles were provided in full
with titles.

\subsubsection{Analysis of Results of Human Sentence Extraction}

As expected with newswire and other journalistic text, many individuals
chose the first sentence.  Although some subjects just took only the first
sentence for each article as a summary, the majority picked several
sentences, usually including the first sentence.  Subjects implicitly
followed the guidelines to pick whole sentences; no readers selected
phrases or sentence fragments.  Subjects indicated that this was not a
difficult task, unlike the segmentation task.

\subsubsection{Use}

To establish the evaluation gold standard, we again applied the majority
method, which resulted in choosing all sentences that were selected by at
least 2 of 3 judges as ``informative''.  The data was used for the
automatic evaluation of an algorithm developed at Columbia, which exploits
both symbolic and statistical techniques.  The sentence extraction
algorithm we have developed uses ranked weighting for information from a
number of well established statistical heuristics from the information
retrieval community, such as TF*IDF, combined with output from term
identification, segmentation, and segment function modules discussed in the
first part of the paper.  Additional weight is given to sentences
containing title words.  Furthermore, several experimental symbolic
techniques were incorporated as factors in the sentence selection weighting
process: such as looking for verbs of communication (Klavans and Kan,
1998, to appear).

An informal examination of the data revealed high level of consistency
among very important sentences, but a lower level of consistency when
important detail was given.  We suspect that the reason may be due to the
equivalency and redundancy of certain sentences.

\subsubsection{Issues} 
\begin{table}[hbt]
\centering
\small
\begin{tabular}{|l|c|c|c|} 
\hline
\hline
& {\bf Article 02} & {\bf Article 18} & {\bf Article 22} \\
& \it{S} = 20 & \it{S} = 20 & \it{S} = 26 \\
\hline
\hline
Subject 1 & 1 & 2 & 1 \\
Subject 2 & 1 & 2 & 1 \\
Subject 3 & 1 & 2 & 1 \\
\hline

\hline
\hline
& {\bf Article 03} & {\bf Article 10} & {\bf Article 11} \\
& \it{S} = 26 & \it{S} = 15 & \it{S} = 17 \\
\hline
\hline
Subject 4 & 4 & 4 & 4 \\
Subject 5 & 3 & 2 & 2 \\
Subject 6 & 1 & 1 & 1 \\
\hline

\end{tabular}
\caption{Verbose and Terse extracters phenomenon in Sentence Extraction
Evaluation Corpora (where {\it S} = number of sentences in article)}
\label{t:table2}
\end{table}

As mentioned in the first section, the project which this resource was
collected for consists of extraction of key sentences from text, and
reformulation of a subset of these sentences into a coherent and concise
summary.  As such, our task is to extract more sentences than would be
explicitly needed for a summary.

The primary challenge in building this resource is analogous to the lumpers
versus splitters difference discussed in Section 2.1.5.  For extraction,
the issue is embodied in the verbose versus terse extractors, i.e.  the
number of sentences selected by subjects had a wide range.  Some subjects
consistently picked very few or just one sentence per article, whereas
others consistently picked many more.  This is shown in Table 2, where for
example, subject 1 picked one or two sentences from each article over 20
sentences or more; whereas both subjects 2 and 3 picked an average of five
sentences from the same article.  Similarly, subject 6 consistently picked
only one sentence, but subject 4 picked four sentences.  This phenomenon,
coupled with the use of a majority method evaluation biases results for
high precision rather than high recall.  Thus, there is a mismatch between
what we asked people to do and what the program was to produce.  We believe
that our compiled resource may be even better suited for an evaluation of a
summarization approach based purely on sentence extraction, although it is
still useful for our evaluation.

\subsubsection{Future Work on the Extraction Resource}

We could compensate for the mismatch in task and algorithm above in two
ways.  One is in the way instructions are given; we could ask subjects to
pick all of the sentences that could be considered of high information
content, or we could give a number of sentences we would like them to pick
for each article.  For the very verbose, we could place an upper bound on
the number of selected sentences.  This could be done simply as some
function of article length, logarithmic or linear.  In the current
collection, we found that some readers thought nearly every sentence was
important, and this affected precision in the final evaluation task.  Some
constraints would push our results towards the more verbose, and eliminate
both the terse subject and the excessively verbose.  Another approach is to
relax the constraints for calculating the gold standard.  As mentioned
above, the majority method in conjunction with the lumpers versus splitters
phenomenon biases results for high precision.  In future work, we will
investigate other methods for culling an evaluation corpus for ``correct''
answers, such as fractional recall and precision (Hatzivassiloglou and
McKeown 93).

\subsubsection{Availability}

The sentence extraction corpora is also publicly available; send any
requests to the first author.  Again, inquiries for the textual data that
the evaluation corpus is based on should be directed to the respective
owners of the materials.
\vspace{-.5cm}

\section{Conclusion}
\vspace{-.2cm}

We have created two corpus resources to be used as a gold standard in the
evaluation of two modules in the analysis stage of a summarization system.
We have discussed several fundamental issues that must be considered in
the effective construction of evaluation resources.
With an increasing number of publicly available
evaluation resources such as these, we contribute to the goals of the
collective sharing of resources and techniques to enable the NLP community
to improve the quality of our future work.

\vspace{-.4cm} 
\section{References}
\vspace{-.4cm}

\begin{list}{}{
\setlength{\rightmargin}{0cm}
\setlength{\leftmargin}{0.35cm}
\setlength{\itemindent}{-0.35cm}
\setlength{\parsep}{0cm}
\setlength{\itemsep}{0.12cm}}

\item Gale, William, Kenneth W. Church and David Yarowsky. (1992).  Estimating upper and lower bounds on the performance of word-sense disambiguation programs.  Proceedings of the ACL.  Newark, Delaware.  pp. 249-256.
\item Hatzivassiloglou, Vasileios and Kathleen R. McKeown. (1993). Towards the Automatic Identification of Adjectival Scales: Clustering Adjectives According to Meaning.  Proceedings of the 31st Annual Meeting of the Association for Computational Linguistics.  Columbus, Ohio, USA.  pp. 172-182.
\item Kan, Min-Yen, Judith L. Klavans, Kathleen R. McKeown (1998)  Linear Segmentation and Segment Relevance.  Columbia University Computer Science Department Technical Report.
\item Klavans, Judith L. and Min-Yen Kan.  (1998, to appear).  The Role of Verbs in Document Analysis.  Proceedings of the 1998 COLING / ACL Conference.  Montr\'{e}al, Qu\'{e}bec, Canada.
\item Marcu, Daniel. (1997)  The Rhetorical Parsing of Unrestricted Natural Language Texts.  Proceedings of the 35th Annual Meeting of the Association Computational Linguistics and 8th Conference of the European Chapter of the Association for Computational Linguistics.  Madrid, Spain.  pp. 96-103.
\item McKeown, Kathleen R. and Dragomir R. Radev. Generating Summaries of Multiple News Articles.  Proceedings, ACM Conference on Research and Development in Information Retrieval SIGIR'95 (Seattle, WA, July 1995).
\item Radev, Dragomir R. and Kathleen R. McKeown. (1997). Building a Generation Knowledge Source using Internet-Accessible Newswire. Proceedings of the 5th Conference on Applied Natural Language Processing. Washington, D.C.

\end{list}


\end{document}